\documentclass{ecai}
\usepackage{times}
\usepackage{graphicx}
\usepackage{latexsym}
\usepackage{amsmath}

\begin{document}

\title{SPCNet: Spatial Preserve and Content-aware \\
Network for Human Pose Estimation}

\author{Yabo Xiao\institute{Beijing University of Posts and Telecommunications, email: [xiaoyabo][wj2718][lvtianqi][fanyiqi][wlr]@bupt.edu.cn. * indicates corresponding author.} \and Dongdong Yu\institute{ ByteDance AI Lab, email: yudongdong@bytedance.com.}  \and Xiaojuan Wang$^{1 *}$ \and Tianqi Lv\footnotemark[1] \and Yiqi Fan\footnotemark[1] \and Lingrui Wu\footnotemark[1] }

\maketitle
\bibliographystyle{ecai}


\begin{abstract}
  Human pose estimation is a fundamental yet challenging task in computer vision. Although deep learning techniques have made great progress in this area, difficult scenarios (e.g., invisible keypoints, occlusions, complex multi-person scenarios,  and abnormal poses) are still not well-handled. To alleviate these issues, we propose a novel \textit{Spatial Preserve and Content-aware Network} (SPCNet), which includes two effective modules: \textit{Dilated Hourglass Module} (DHM) and \textit{Selective Information Module} (SIM). By using the \textit{Dilated Hourglass Module}, we can preserve the spatial resolution along with large receptive field. Similar to Hourglass Network, we stack the DHMs to get the multi-stage and multi-scale information. Then, a \textit{Selective Information Module} is designed to select relatively important features from different levels under a sufficient consideration of spatial content-aware mechanism and thus considerably improves the performance. Extensive experiments on MPII, LSP and FLIC human pose estimation benchmarks demonstrate the effectiveness of our network. In particular, we exceed previous methods and achieve the state-of-the-art performance on three aforementioned benchmark datasets.
  
\end{abstract}

\section{Introduction}

Human pose estimation aims to locate the person parts, such as keypoints on the arms, legs and face. It is a fundamental yet challenging task in computer vision, which plays an important role in many high-level vision tasks like activity understanding \cite{chunyu2013approach} and human re-identification \cite{zheng2017pose}.

With the development of Convolutional Neural Network \cite{tompson2014joint,toshev2014deeppose,wei2016convolutional,tompson2015efficient,newell2016stacked}, great progress has been achieved in human pose estimation. For example, in \cite{newell2016stacked}, Newell first proposes the hourglass network to predict the human keypoints. It shows that the repeated bottom-up, top-down processing and the intermediate supervision are critical to improving the estimation performance. Yang et al.\cite{yang2017learning} design a Pyramid Residual Module to explicitly learn convolutional filters for building feature pyramids and enhance the robustness of keypoint estimation against scale variations of visual patterns. Although the stacked hourglass network \cite{newell2016stacked} and its variants \cite{ke2018multi,yang2017learning,chu2017multi,zhang2019human} have achieved significant performance, it is still an open problem to achieve accurate localizing results due to the occluded keypoints, overlapped limbs, and abnormal poses. 

\begin{figure}
\begin{center}
\includegraphics[height=0.6\columnwidth,width=1.0\columnwidth]{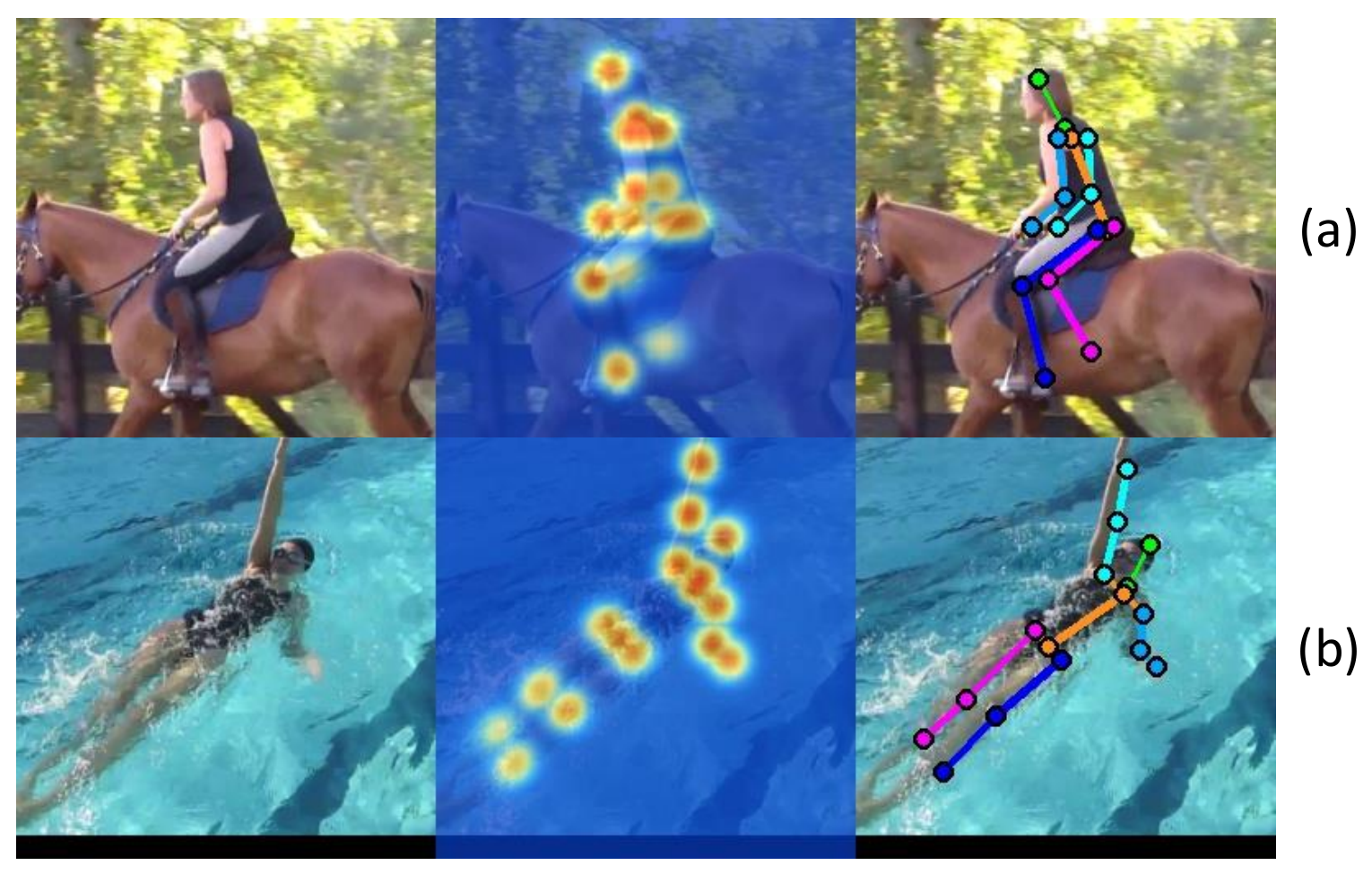}
\end{center}
\caption{Complex scenarios predictions on MPII Human Pose test dataset by our proposed network. (a) A person with invisible limbs followed by predicted heatmap and skeleton. (b) A person with blurred and abnormal pose followed by predicted heatmap and skeleton.}
\label{fig:image1}
\end{figure}

\begin{figure*}
\begin{center}
\includegraphics[height=0.9\columnwidth,width=1.8\columnwidth]{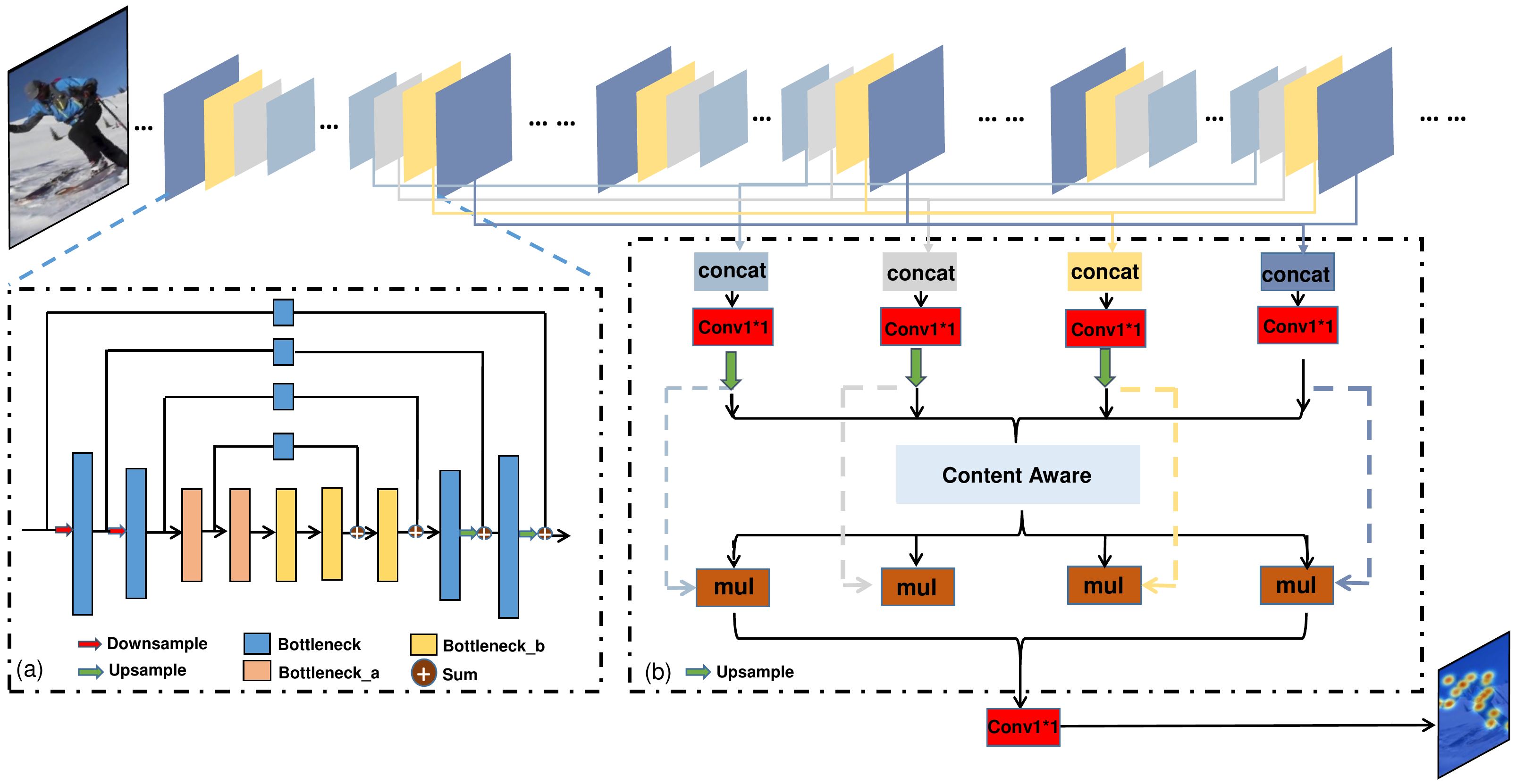}
\end{center}
\caption{ Overview of SPCNet. ~(a) The structure of \textit{Dilated Hourglass Module}(DHM). ~(b) The diagram of \textit{Selective Information Module}(SIM). We stack 8 DHMs to capture multi-level features, then the SIM adaptively fuses the multi-level features under a sufficient consideration of spatial content-aware mechanism.}
\label{fig:image2}
\end{figure*}

To locate keypoints accurately, a model has to take the high spatial resolution information and multi-level information (e.g., multi-stage information and multi-scale information) into account. For example, the right ankle and right knee of the woman in Figure \ref{fig:image1}(a) are invisible keypoints, high-level feature maps with large receptive field are needed to infer such kind of keypoints. However, the high spatial resolution information can provide detailed features, which is useful for refining the positions of the other visible joints. Another example, the human body in Figure \ref{fig:image1}(b) is blurred with abnormal pose, multi-level features should be extracted and fused effectively and sufficiently for the blurred and abnormal pose estimation. According to the above analysis, a \textit{Dilated Hourglass Module} is proposed to preserve high spatial resolution information along with large respective field. Moreover, a \textit{Selective Information Module} is designed to fuse the features of different levels under a sufficient consideration of spatial content-aware mechanism. Based on the two prominent components, we propose an efficient Spatial Preserve and Content-aware Network (SPCNet) for human pose estimation, as shown in Figure \ref{fig:image2}.

For preserving high spatial resolution information along with large respective field, we exploit the dilated convolution operation in the hourglass network. Classic hourglass network adopts large downsampling factor to obtain large receptive field. It is good for estimating the occluded and invisible keypoints, but compromises the location ability. Different from the hourglass network, we substitute the hourglass module with our proposed \textit{Dilated Hourglass Module} (DHM) to obtain large receptive field and avoid the reduction of spatial dimension caused by the subsampling operation. As shown in Figure \ref{fig:image2}(a), the spatial size is fixed after only two subsampling operations. Then the dilated bottlenecks are applied to keep the spatial resolution, which can also efficiently capture semantic information and maintain detailed features.
 
Nowadays, there is an increasing interest in designing networks with attention mechanism. Recent works focus on the spatial attention, channel attention, and non-local attention in the single-level information, and prove the effectiveness of the attention module. However, little attention has been paid to adaptively fuse multi-level information under visual attention mechanism. In our paper, a \textit{Selective Information Module} (SIM) is proposed to effectively fuse the different levels of features under the attention mechanism. As described in Figure \ref{fig:image2}(b), we adaptively assemble the multi-level features via a pixel-wise weighted summation in spatial dimension, where the pixel-wise weights are produced by attention-based method under a sufficient consideration of spatial content-aware mechanism. At different spatial position, multi-level features are fused in different proportions according to the diversity of local region content.

In this paper, we first gather up the multi-stage information and multi-scale information from the decoder  layers  of  each  \textit{Dilated  Hourglass  Module} to compose four high-dimensional feature maps of different levels. Second, we use the proposed \textit{Selective Information Module} to effectively fuse the four different levels information for predicting the human body keypoints. Then, we evaluate our proposed method on MPII Human Pose dataset \cite{andriluka20142d} , LSP dataset \cite{johnson2010clustered} and FLIC dataset\cite{sapp2013modec}, ablation studies demonstrate the effectiveness of the \textit{Dilated Hourglass Module} and \textit{Selective Information Module}. In particular, our method exceeds prior methods and achieves the state-of-the-art performance. 

In summary, there are three contributions in our paper:
\begin{itemize}
\item We explore a novel \textit{Dilated Hourglass Module} which employs the dilated bottlenecks to preserve high spatial resolution and obtain large receptive field.

\item We propose an effective feature fusion method called \textit{Selective Information Module}, which is able to adaptively assemble the multi-level spatial information.  

\item Our proposed network outperforms the state-of-the-art methods on MPII Human Pose dataset, LSP dataset and FLIC dataset.
\end{itemize}

The rest of this paper is organized as follows. In Section \ref{sec2}, we review some articles related to our work. In Section \ref{sec3}, we present the main idea of our SPCNet work. Then ablation studies are performed to measure the effects of different parts of our system, and the experimental
results are reported in Section \ref{sec4}, followed by a conclusion in Section \ref{sec5}.

\section{Related Work}
\label{sec2}
In this section, we review three parts related to our method: human pose estimation, dilated convolution and attention mechanism.

\noindent{\bf Human pose estimation.}  There are many application scenarios for human pose estimation, such as activity understanding \cite{chunyu2013approach}, human re-identification \cite{zheng2017pose}. Pictorial structures \cite{andriluka2009pictorial} or graphical models \cite{chen2014articulated} as representative of the traditional methods are used to deal with pose estimation problems. However, these methods predict positions of keypoints rely on hand-generated features, which are susceptible to difficult issues such as occlusion. Recently, deep convolutional networks surpass traditional methods and achieve the state-of-the-art results in pose estimation. DeepPose \cite{toshev2014deeppose} uses deep learning method for pose estimation which directly regresses the keypoints’ coordinates by multi-stage refinement for the first time. Methods of ~\cite{newell2016stacked,wei2016convolutional,chen2018cascaded,Cao2017Realtime,Sun2019Deep} use fully convolutional neural network to regress the Gaussian heatmap and infer the human keypoint coordinates by using the Gaussian peaks. These methods can produce high quality representation. In \cite{newell2016stacked}, Newell first proposes the hourglass network to predict the human keypoints. It shows that the repeated bottom-up, top-down processing and the intermediate supervision are critical to improving the estimation performance. Pyranet\cite{yang2017learning} is a variant of stacked hourglass network that designs the pyramid residual module to enhance the invariance in scales of deep convolutional neural network. 

\noindent{\bf Dilated convolution.}  Recently, lots of approaches with dilated convolution have achieved high performance on different benchmarks of semantic segmentation~\cite{chen2017rethinking} and object detection~\cite{Li2019Scale}. DeepLab \cite{chen2017rethinking} designs atrous spatial pyramid pooling (ASPP) that applies dilated convolution with various dilation rates on multiple parallel branches to capture detailed information and context information(multi-level informations). One key advantage is that it can effectively enlarge receptive field size to incorporate context without introducing extra parameters or computation cost. In parallel, large receptive field is also needed for the hard keypoints' prediction. Motivated by this, we propose a novel bottom-up and top-down hourglass module decorated with the dilated convolution. In this way, we can obtain large receptive field and maintain high spatial resolution simultaneously, both of which are critical to the human pose estimation task.

\noindent{\bf Attention mechanism.}  Visual attention has achieved great success in various tasks, such as image classification, human pose estimation and image segmentation. SENet\cite{hu2018squeeze} proposes a "Squeeze-and-Excitation" block to recalibrate channel-wise feature by using channel attention operation. In~\cite{su2019multi}, Su et al. design a Spatial Channel-wise Attention Residual Bottleneck to enhance the feature responses both in the spatial and channel-wise context. The above methods only study the spatial attention and channel attention concentrating on the single-level information. However, little attention has been paid to adaptively fuse multi-level information under visual attention mechanism. Our work is inspired by the spatial-attention approaches and we adaptively assemble the multi-level spatial features by fusing the information of different levels via a pixel-wise weighted summation, where the weighted parameters are learned through the spatial attention mechanism.

\section{Method}
\label{sec3}

In this section, we propose a novel \textit{Spatial Preserve and Content-aware Network} (SPCNet) to preserve spatial resolution information and select relatively important features from different levels according to the local part content under the spatial attention mechanism. An overview of the proposed SPCNet is illustrated in Figure \ref{fig:image2}. First, we briefly review the stacked hourglass network \cite{newell2016stacked}. Then, we introduce the structure of \textit{Dilated Hourglass Module} (DHM) and \textit{Selective Information Module} (SIM) in detail. Finally, the complete network architecture of SPCNet is presented as well as the training and inference processing details.

\subsection{Revisiting Stacked Hourglass Network}
Stacked hourglass network \cite{newell2016stacked} is a classic approach for locating body keypoints from RGB images. The hourglass unit performs bottom-up process by subsampling the feature maps, and then conducts symmetric top-down process by upsampling the feature maps with the combination of higher resolution features from bottom layers to generate the high resolution heatmaps. Then, the hourglass units are stacked to build the stacked hourglass network. Each hourglass unit is supervised with the ground-truth heatmap.

\begin{figure}
\begin{center}
\includegraphics[height=0.63\columnwidth,width=0.9\columnwidth]{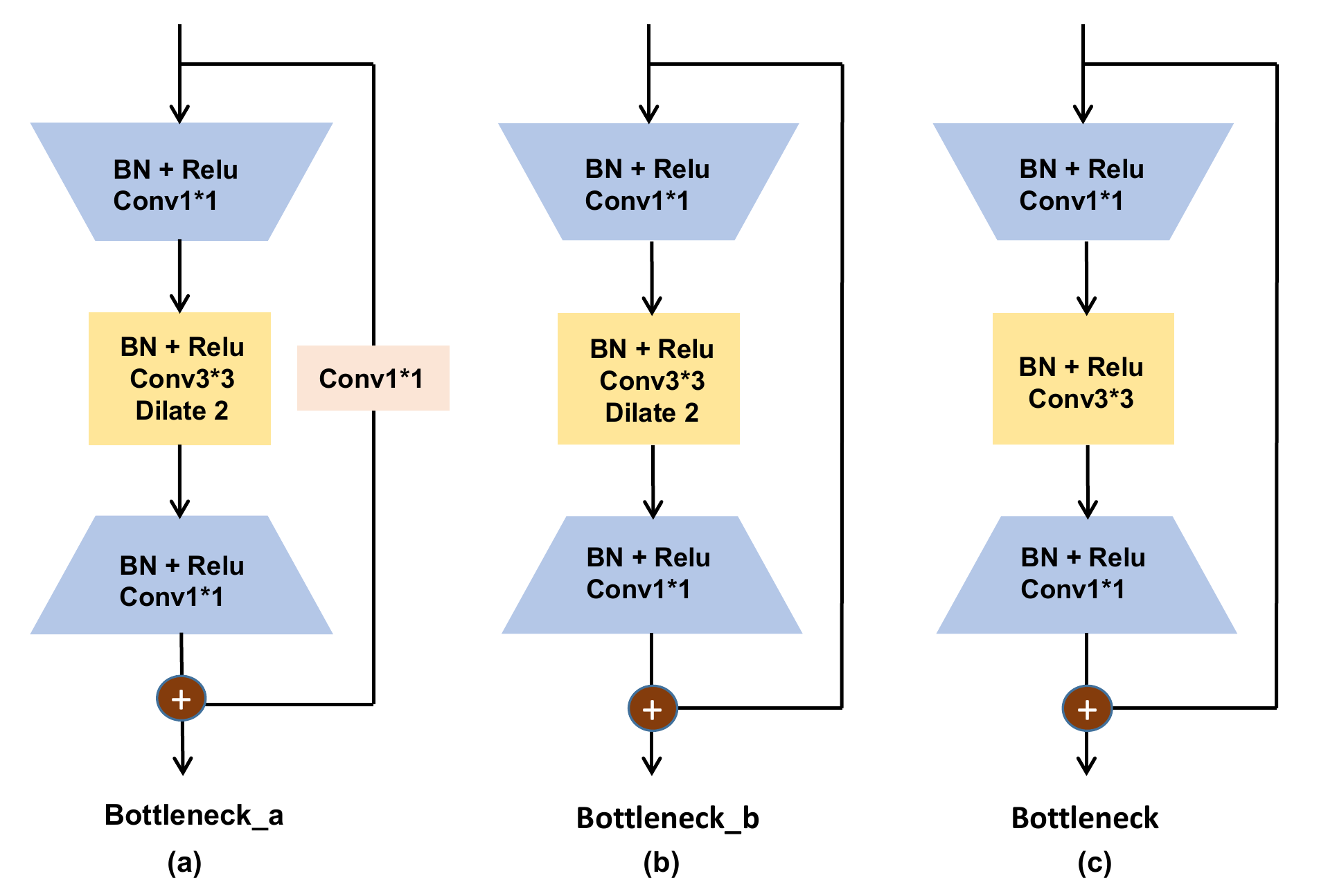}
\end{center}
\caption{(a)(b) Detailed structure of two different dilated bottlenecks used in \textit{Dilated Hourglass Module}. (c) Conventional bottleneck.}
\label{fig:image3}
\end{figure}

\subsection{Dilated Hourglass Module}
\noindent{\bf Motivation}~~In the task of single-person pose estimation, most of the modern methods tackle it as a dense regression issue. Large downsampling factor in encoding process brings large effective receptive field, which is beneficial to the inference of occluded, twisted and overlapped limbs. However, it will reduce the spatial resolution which is critically important for refining the location of human body keypoints. Therefore, the trade-off between the global feature and the detail feature should be taken into account. Motivated by this, we propose a novel \textit{Dilated Hourglass Module} not only to maintain the high resolution spatial structure on human body but also obtain large receptive field, which is shown in Figure \ref{fig:image2}(a).

In the bottom-up way of the original hourglass unit, 
it conducts 4 downsampling operations to reduce the spatial dimension from $64\times64$ to $4\times4$ pixels for obtaining large receptive field.  The lowest resolution feature map contains high-level semantic information which is critical to the hard keypoints' location, while it compromises the spatial resolution information which can provide detailed information for refining the positions of keypoints. Aiming at leveraging both the abundant context information and high spatial resolution information, we proposed a \textit{Dilated Hourglass Module} (DHM) in our paper.

On one hand, the DHM uses only two downsampling operations to maintain the high spatial dimension which is 1/4 of the input resolution and fix the spatial resolution as 4x downsampling after two downsampling operations to mitigate the aforementioned issue. On the other hand, in order to get large receptive field, we introduce $3\times3$ convolutional layer with dilation rate $R$  to replace the conventional $3\times3$ convolutional layer of original residual block.  The dilated convolution  process is formulated as follows:

\begin{equation}\label{eq:3}
out[i,j] = \sum_{m=0}^{K-1}\sum_{n=0}^{K-1}(inp[i+R*m,j+R*n]*W[m,n]),
\end{equation}

where $out[i,j]$ is the value of output feature map at spatial position$[i,j]$, ~$inp$ is the input feature map, $[m,n]$ represents the position index of the dilated convolutional kernel $W$, $K$ refers to the kernel size and $R$ corresponds to the dilation rate. For instance, the kernel size of $3\times3$ convolution filter with dilation rate R could be considered as $3 + 2\times({R}  -  1)$. In our paper, there are three kinds of residual bottlenecks in our proposed DHM: dilated residual blocks (Bottleneck\_a and Bottleneck\_b), conventional residual block (Bottleneck) as shown in Figure \ref{fig:image3}. R is experimentally set to 2 for Bottleneck\_a and Bottleneck\_b.

\begin{figure*}
\begin{center}
\includegraphics[width=1.8\columnwidth]{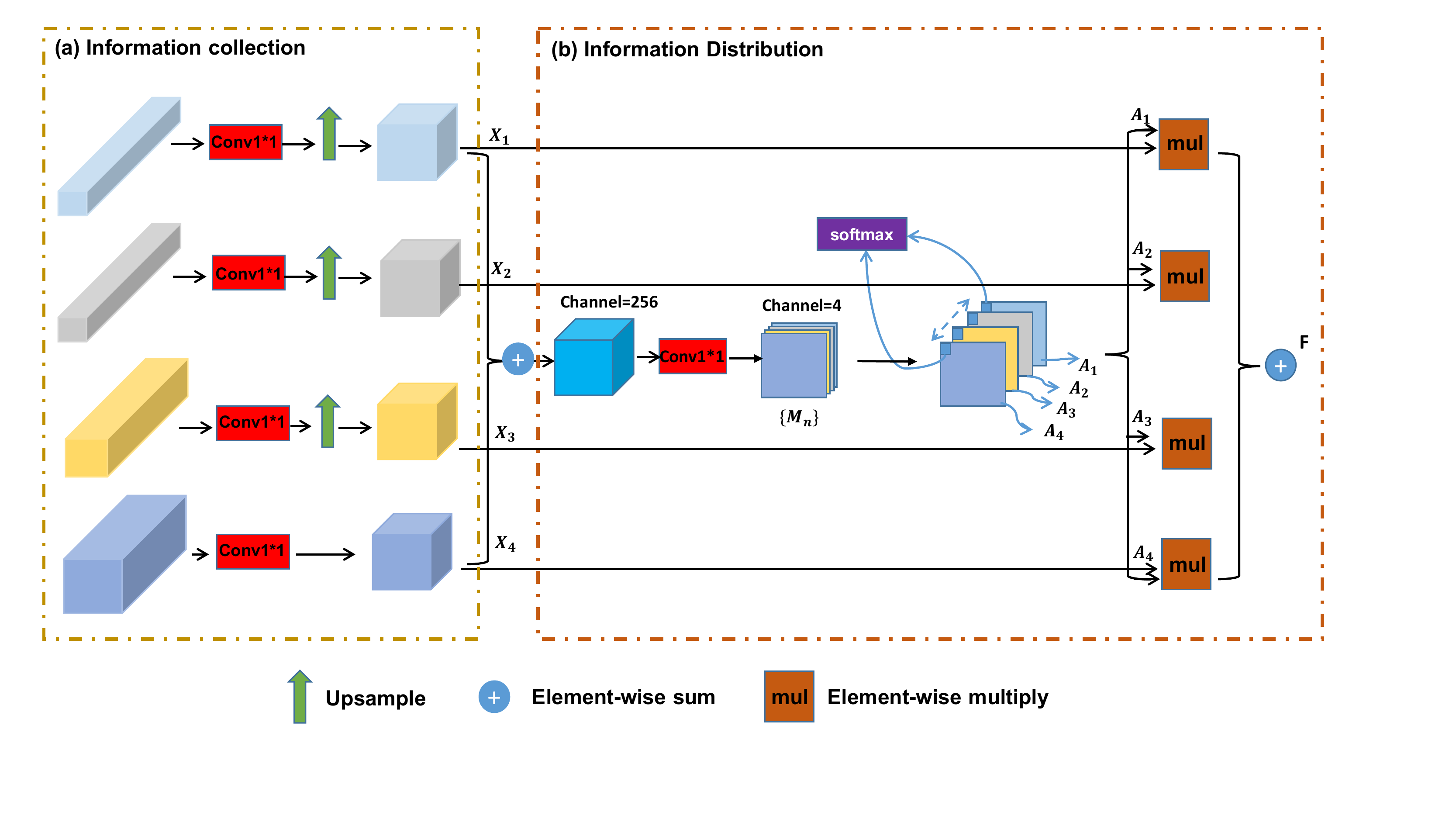}
\end{center}
\caption{The overall structure of \textit{Selective Information Module}(SIM), the SIM consists of two steps: ~(a) {\bf Information Collection}: $\mathbf{X}_{1}\sim\mathbf{X}_{4}$ are multiple receptive field features with equal dimension extracted by Information Collection process. ~(b) {\bf Information Distribution}: $\left\{\mathbf{M}_{n}\right\}$ is the squeezed feature with 4 channels before softmax operation. $\mathbf{A}_{1}\sim\mathbf{A}_{4}$ are pixel-wise weights normalized along the channel via a softmax layer.}
\label{fig:image4}
\end{figure*}

\subsection{Selective Information Module}\label{3.3}

\noindent{\bf Motivation} High resolution feature maps consist of more details which can precisely localize the keypoints, but may fail to recognize the twisted or overlapped keypoints due to small receptive field. Low resolution representations contain more semantic context which can handle the difficult scenarios (e.g., invisible keypoints, twisted pose). How to efficiently aggregate the multi-level features is still a challenging issue for the pose estimation.  In general, element-wise summation and channel-wise concatenation are most commonly used methods to fuse multi-level features. Element-wise summation equally aggregate the features from multiple levels, which is not a learnable process. Despite the channel-wise concatenation followed by $1\times1$ convolution can be considered as a learnable process, while it applies the same convolutional kernel across the whole feature map regardless of the content of local region. However, the importance of each level should be treated different in intuition. We consider that different local regions require multi-level features in different proportions according to the diversity of local region content information.

Against the aforementioned problem, we design \textit{Selective Information Module} (SIM) to adaptively assemble the spatial features from multiple levels, as illustrated in Figure \ref{fig:image4} . There are two main steps in our proposed SIM: \textbf{Information Collection} and \textbf{Information Distribution}. With the Information Collection, we can get multi-level feature information from the stacked DHMs. Through the Information Distribution, the multi-level feature information can be adaptively fused. In detail, our proposed module fuses the multi-level features via a weighted summation at each pixel position, where the weights are generated by trainable process. We experimentally verified  our proposed module effectively fuse the spatial detail information and context information.

\textbf{Information Collection:} We first extract the multi-level features(i.e., 16, 16, 32, 64 pixels) from each deconv layer of the \textit{Dilated Hourglass Module} (DHM). Second, features with equivalent scales in eight stacked DHMs are concatenated to obtain four high-dimensional feature maps. Then, we reduce the channels of the four high-dimensional feature maps by $1\times1$ convolutional layers which reduces the number of channels from 2048 to 256 followed by the batch normalization and ReLU in sequence. Next, we upsample multi-level features(i.e., 16, 16, 32, 64 pixels) to $64\times64$ respectively and denote the features with multiple receptive field as $\mathbf{X}_{1}\sim\mathbf{X}_{4}$ (with the same resolution of $64\times64$ and the same channel of 256 ). 

\textbf{Information Distribution:} We first simply fuse $\mathbf{X}_{1}\sim\mathbf{X}_{4}$ via element-wise combination which is considered as information aggregation from features of multiple levels. Then we employ $1\times1$  convolutional filters to conduct the channel squeeze and compress the channels to 4. Here we consider the squeezed feature $\mathbf{M}=\left\{ \mathbf{M}_{n}\right\}_{n=1}^4$, where $\mathbf{M}_{n} = \{{M}_{n}^{1,1}, {M}_{n}^{1,2},..., {M}_{n}^{i,j},...,{M}_{n}^{H,W}\}$ corresponding to the  feature map of single channel. Index $(i,j)$ represents the pixel position. After that, a softmax operation is conducted across channels to rescale activations and then rescaled activations are sliced along the channel dimension to generate pixel-wise adaptive weights $\mathbf{A}=\left\{ \mathbf{A}_{n}\right\}_{n=1}^4$ corresponding to the multi-level features(i.e., $\mathbf{X}_{1} \sim \mathbf{X}_{4}$). The process above are formulated as follows:
\begin{equation}
   {A}_{n}^{i,j} = \frac{exp({M}_{n}^{i,j})}{\sum_{k=1}^{4}{exp({M}_{k}^{i,j})}} ,         
\end{equation}

\begin{equation}\label{eq:2}
 \sum_{n=1}^{4}({A}_{n}^{i,j})=1,
\end{equation}
in which, ${A}_{n}^{i,j}$ represents relative importance of multi-level features $\mathbf{X}_{n}$ at  pixel position $(i,j)$. ${A}_{1}^{i,j} \sim {A}_{4}^{i,j}$ are the learned weights for ${X}_{1}^{i,j} \sim {X}_{4}^{i,j}$ respectively. The addition of $\mathbf{A}$ along the channel dimension is normalized to 1 via softmax operation for each spatial position $(i,j)$. Finally the assemble for features with multiple receptive field  is defined as:

\begin{equation}\label{eq:4}
  \mathbf{F} =   \sum_{n=1}^{4}(\mathbf{A}_{n} \ast \mathbf{X}_{n}),
\end{equation}
where {\bf{F}} is the final fused feature for predicting the human keypoints, $\ast$ means the pixel-wise multiplication.

\subsection{Network Architecture, Training and Inference}

\noindent{\textbf{Network Architecture.}} With the \textit{Dilated Hourglass Module} and \textit{Selective Information Module}, we propose a novel \text{Spatial Preserve and Content-aware Network} (SPCNet) for human pose estimation as illustrated in Figure \ref{fig:image2}. First, we propose a \textit{Dilated Hourglass Module} to preserve the spatial resolution along with large receptive field. Similar to Hourglass Network, we stack the DHMs to get the multi-stage and multi-scale information. Then, a \textit{Selective Information Module} is designed to select relatively important features from different levels under a sufficient consideration of spatial content-aware mechanism. Finally, we use the selected information by the SIM to predict the human body keypoints. In our paper, we find that the two proposed modules are complementary to each other for higher performance on keypoint localization.

\noindent{\textbf{Network Training.}}~~We use score maps to represent the ground-truth heatmaps of human body keypoints.  Denote the ground-truth positions by $\mathbf{C}=\left\{ \mathbf{C}_n \right\}_{n=1}^N$, $N$ is the number of human body keypoints. $\mathbf{C}_n$ is the coordinates of the $n$th keypoint. In this paper, we use a Gaussian distribution with mean $\mathbf{C}_n$ and variance $ \delta $ to represent the ground-truth heatmap $ \mathbf{Y}_n $ as follows:
\begin{equation}
\mathbf{Y}_n \sim N(\mathbf{C}_n,\delta).   
\end{equation}

A squared error loss function is applied to minimize the loss between the predicted score maps $\mathbf{\hat{Y}}_n$ (each \textit{Dilated Hourglass Module} and the \textit{Selective Information Module}) and the ground-truth heatmaps:
\begin{equation}
 J = \frac{1}{2}\sum_{i = 1}^{9}\sum_{k = 1} ^K \sum_{n = 1} ^N(\mathbf{Y}_n - \mathbf{\hat{Y}}_n)^2,
\end{equation}
$K$ is the number of samples. In our paper, there are 8 auxiliary losses from the stacked Dilated Hourglass Module and 1 supervision loss for the \textit{Selective Information Module}.

\noindent{\textbf{Network Inference.}}~~During inference, we obtain the predicted body joint locations from the predicted score maps generated from the \textit{Selective Information Module} by taking the locations with the maximum score as follows:
\begin{equation}
\mathbf{\hat{C}}_n=\mathop{\arg\max} \mathbf{\hat{Y}}_n ,  n= 1,2,3,..., N.
\end{equation}

\section{Experiments and Analysis}
\label{sec4}
In this section, we first briefly introduce the datasets, evaluation metrics and implementation details in \ref{subsection1}. Next, we conduct comprehensive ablation study to reveal the effectiveness of our proposed modules in \ref{subsection2}. Finally, we compare our results with the prior state-of-the-art results on MPII Human Pose dataset \cite{andriluka20142d}, LSP dataset \cite{johnson2010clustered} and FLIC dataset\cite{sapp2013modec}. 

\subsection{Experimental Setup}\label{subsection1}

{\bf Datasets and Evaluation Metrics.} We evaluate the performance of our network on three benchmark datasets mentioned above. The MPII Human Pose dataset is composed of around 25K images containing over 40K samples with annotated body joints. 28K samples are used for training, and the remaining 12K are used for testing. The LSP dataset and its extended training dataset includes 12K sports images with annotations (11K images are used for training and 1K images are used for testing). The FLIC dataset consists of 5003 samples(3987 for training, 1016 for testing). The evaluation is conducted using Percentage of Correct Keypoints (PCK) \cite{andriluka20142d} metric which shows the percentage of detections that fall within a normalized distance of the ground truth. For the MPII Human Pose evaluation, we use the modified PCK measure that uses a fraction of head size as the normalized factor(denoted as PCKh\cite{andriluka20142d}). For the LSP and FLIC evaluation, we use PCK as previous researches~\cite{yang2017learning,tang2018deeply}. 

\noindent{\bf Data Augmentation.} During training, we use random rotation, random flip, and random scaling. The rotation range is (-60, 60), the flip probability is 0.5 and the scale range is (0.75, 1.25). Each input image is cropped around the target person according to the annotated body center and scale, and then resized to $256\times256$ pixels.

\noindent{\bf Implementation Details.} We train our proposed network using RMSProp \cite{tieleman2012lecture} optimizer with a mini-batch size of 48 (12 per GPU) for 170 epochs on a workstation with four 12GB NVIDIA TITAN XP GPUs. The initial learning rate of 1e-3 and is dropped by the factor of 10 at the 120th and the 150th epoch. All codes are implemented with Pytorch. A Mean Squared Error (MSE) loss is applied to compute the loss between the predicted heatmap and the ground-truth heatmap. Testing results are produced from six-scale image pyramids with flipping.

\subsection{Ablative Analysis}\label{subsection2}

In this subsection, we conduct ablation experiments on the validation set of the MPII Human Pose to explore the effectiveness of the proposed \textit{Dilated Hourglass Module} (DHM) and \textit{Selective Information Module} (SIM). We define the 8-stack hourglass network as our baseline network. Based on the hourglass network, we first explore each proposed component and then conduct comprehensive analysis for the impact of each module (i.e. DHM and SIM) for the whole network by comparing the PCKh score.

\noindent{\bf Effect of Dilated Hourglass Module.} In this experiment, we investigate the effect of our proposed \textit{Dilated Hourglass Module} and the influence of dilation rate R. The original hourglass module is replaced by the DHM with various dilation rate to conduct a series of experiments. As shown in Table \ref{tab:dilation}, we achieve the best performance when R is set to 2. Compared with the baseline network, The PCKh score is improved from 88.9\% to 89.7\% by using \textit{Dilated Hourglass Module} with dilation rate 2, which is an obvious improvement. It experimentally confirmed that using the dilated convolution can preserve more spatial information, which is benefit for refining the positions of joints.

\begin{table}
\renewcommand\arraystretch{1.0}
\begin{center}
\caption{Ablation experiments about \textit{Dilated Hourglass Module} on MPII validation dataset (PCKh@0.5).}
\label{tab:dilation}
\resizebox{1.0\columnwidth}{!}{
\begin{tabular}{lcccccccc}
\hline
 Methods       &  Head & Sho. & Elb. & Wri. & Hip & Knee& Ank.& Total  \\
\hline
 Baseline    & 96.7 & 95.8 & 89.9 &84.9 & 88.8& 85.0 & 80.6 &88.9   \\
 
DHM,R=1   &{\bf96.9} &95.6 &90.0 &85.3 &  89.3 &84.8 &81.0 & 89.0\\
DHM,R=2  & 96.6  & {\bf96.1} & {\bf90.5} & 85.4 & {\bf89.7} & {\bf86.4} & {\bf82.3}& {\bf89.7} \\ 
DHM,R=3 &{\bf96.9} &95.7 &90.1 &{\bf85.6}  & 89.4 &85.8 &81.9 &89.4 \\
  
\hline
\end{tabular}
}
\end{center}
\end{table}

\begin{table}
\begin{center}
\caption{Ablation experiments about \text{Selective Information Module} on MPII validation dataset (PCKh@0.5).}
\label{tab:content-aware}
\resizebox{1.0\columnwidth}{!}{
\begin{tabular}{lcccccccc}
\hline
 Methods       &  Head & Sho. & Elb. & Wri. & Hip & Knee& Ank.& Total  \\
\hline
 Baseline    & 96.7 & 95.8 & 89.9 &84.9 & 88.8& 85.0 & 80.6 &88.9   \\
 Baseline+IC+Sum    & {\bf96.8}  & 95.6 & 90.1 & 84.8 & 89.1 &85.0 & 81.6& 89.1 \\  
 Baseline+IC+Concat & 96.5  & 95.9 & 90.2 & 85.3 & 89.1 &85.3 & 82.1& 89.3 \\ 
 Baseline+SIM(IC+ID)   & {\bf96.8}  & {\bf96.0} & {\bf90.5} & {\bf85.6} & {\bf89.5} &{\bf86.1} & {\bf82.5}& {\bf89.6} \\ 
\hline
\end{tabular}
}
\end{center}
\end{table}



\noindent{\bf Effect of Selective Information Module.} There are two main steps in \textit{Selective Information Module}(SIM): Information Collection(IC) and Information Distribution(ID). To explore the effectiveness of \textit{Selective Information Module}, we first get multi-level features(i.e.,$\mathbf{X}_{1}\sim\mathbf{X}_{4}$) from the stacked DHMs through the Information Collection process, as described in Section~\ref{3.3}, and then conduct a series of experiments among three feature fusion methods: element-wise summation, channel-wise concatenation and Information Distribution process in our proposed SIM. In detail, for element-wise summation, we sum the multi-level features $\mathbf{X}_{1}\sim\mathbf{X}_{4}$ directly; for channel-wise concatenation, we first concatenate the $\mathbf{X}_{1}\sim\mathbf{X}_{4}$ along the channel dimension and then add an extra $1\times1$ convolutional layer after concatenate operation to generate fusion feature by compressing the channel to 256. As shown in Table \ref{tab:content-aware}, we observe that Information Collection process combined with any of the three feature fusion methods can improve the performance of baseline network. Furthermore, the Information Distribution process achieves the best PCKh score among these three feature fusion methods on MPII validation set. We obtain 0.5\% and 0.3\% improvement by replacing the element-wise summation and channel-wise concatenation with the Information Distribution process respectively and achieve 0.7\% improvement by adding the SIM to the original hourglass network, which demonstrate the superior performance of our \textit{Selective Information Module} over the other feature fusion methods.

\begin{table}
\begin{center}
\caption{Ablation experiments about each proposed Module on MPII validation dataset (PCKh@0.5).}
\label{tab:each}
\resizebox{1.0\columnwidth}{!}{
\begin{tabular}{lccccccccc}
\hline
 DHM & SIM   &  Head & Sho. & Elb. & Wri. & Hip & Knee& Ank.& Total  \\
\hline
 $\times$ & $\times$   & 96.7 & 95.8 & 89.9 &84.9 & 88.8& 85.0 & 80.6 &88.9   \\
 $\surd$ &  $\times$  & 96.6  & 96.1 & 90.5 & 85.4 & 89.7 &{\bf86.4} & 82.3& 89.7 \\ 
$\times$  & $\surd$   & 96.8  & 96.0 & 90.5 & 85.6 & 89.5 &86.1 & {\bf82.5}& 89.6 \\ 
 $\surd$ & $\surd$  &  {\bf96.9} & {\bf96.4} & {\bf90.9} & {\bf86.3} & {\bf89.8} &{\bf86.4} & {\bf82.5}& {\bf90.0} \\ 
\hline
\end{tabular}
}
\end{center}
\end{table}

\begin{table}
\begin{center}
{\caption{Performance comparisons on the MPII validation dataset(PCKh@0.5).}\label{tab:six}}
\resizebox{1.0\columnwidth}{!}{
\begin{tabular}{lcccccccc}
\hline
 Methods &  Head & Sho. & Elb. & Wri. & Hip & Knee& Ank.& Total  \\
 
\hline
\multicolumn{9}{c}{single-scale}\\
\hline
Newell et al.~\cite{newell2016stacked} &96.7& 95.8 &89.9 & 84.9 &88.8& 85.0& 80.6& 88.9 \\
 \textbf{Ours(SPCNet)} &  {\bf96.9} & {\bf96.4} & {\bf90.9} & {\bf86.3} & {\bf89.8} &{\bf86.4} & {\bf82.5}& {\bf90.0} \\
\hline
\multicolumn{9}{c}{single-scale with horizon flip}\\
\hline
Newell et al.~\cite{newell2016stacked} & 96.8& 96.0& 90.6 &85.9& 89.8& 86.1& 81.1& 89.5 \\
Yang et al.~\cite{yang2017learning}& 96.8& 96.0& 90.4& 86.0& 89.5& 85.2& 82.3& 89.6  \\
Tang et al.~\cite{tang2018deeply} &95.6& 95.9& 90.7& 86.5& 89.9 &86.6& 82.5& 89.8\\
SimpleBaseline~\cite{xiao2018simple} &97.0 &95.9 &90.3& 85.0 &89.2 &85.3& 81.3& 89.6\\
HRNet~\cite{Sun2019Deep}&{\bf97.1}& 95.9 &90.3& 86.4& 89.1& {\bf87.1}& 83.3& 90.3\\
\textbf{Ours(SPCNet)} & {\bf97.1}&{\bf96.4} &{\bf91.3} & {\bf87.0}& {\bf90.0}& {\bf87.1}& {\bf83.8} & {\bf90.5}  \\
\hline
\multicolumn{9}{c}{multi-scale with horizon flip}\\
\hline
Newell et al.~\cite{newell2016stacked} & 97.1& 96.1& 90.8 &86.2& 89.9 &85.9 &83.5& 90.0\\
Yang et al.~\cite{yang2017learning} &97.4 &96.2& 91.1 &86.9& 90.1& 86.0& 83.9 &90.3 \\
Tang et al.~\cite{tang2018deeply}&97.4 &96.2 &91.0 &86.9 &90.6& 86.8 &84.5& 90.5 \\
SimpleBaseline~\cite{xiao2018simple} &97.5& 96.1& 90.5& 85.4& 90.1& 85.7& 82.3& 90.1\\
HRNet~\cite{Sun2019Deep} &97.7& 96.3& 90.9& 86.7& 89.7& 87.4& 84.1& 90.8\\
\textbf{Ours(SPCNet)}  &{\bf97.8} & {\bf96.6} & {\bf91.9} &  {\bf87.5}  & {\bf90.7} &  {\bf87.5} & {\bf84.5}  & {\bf91.1}\\ 
\hline
\end{tabular}
}
\end{center}
\end{table}

\begin{figure}
\begin{center}
\includegraphics[height=0.8\columnwidth,width=1\columnwidth]{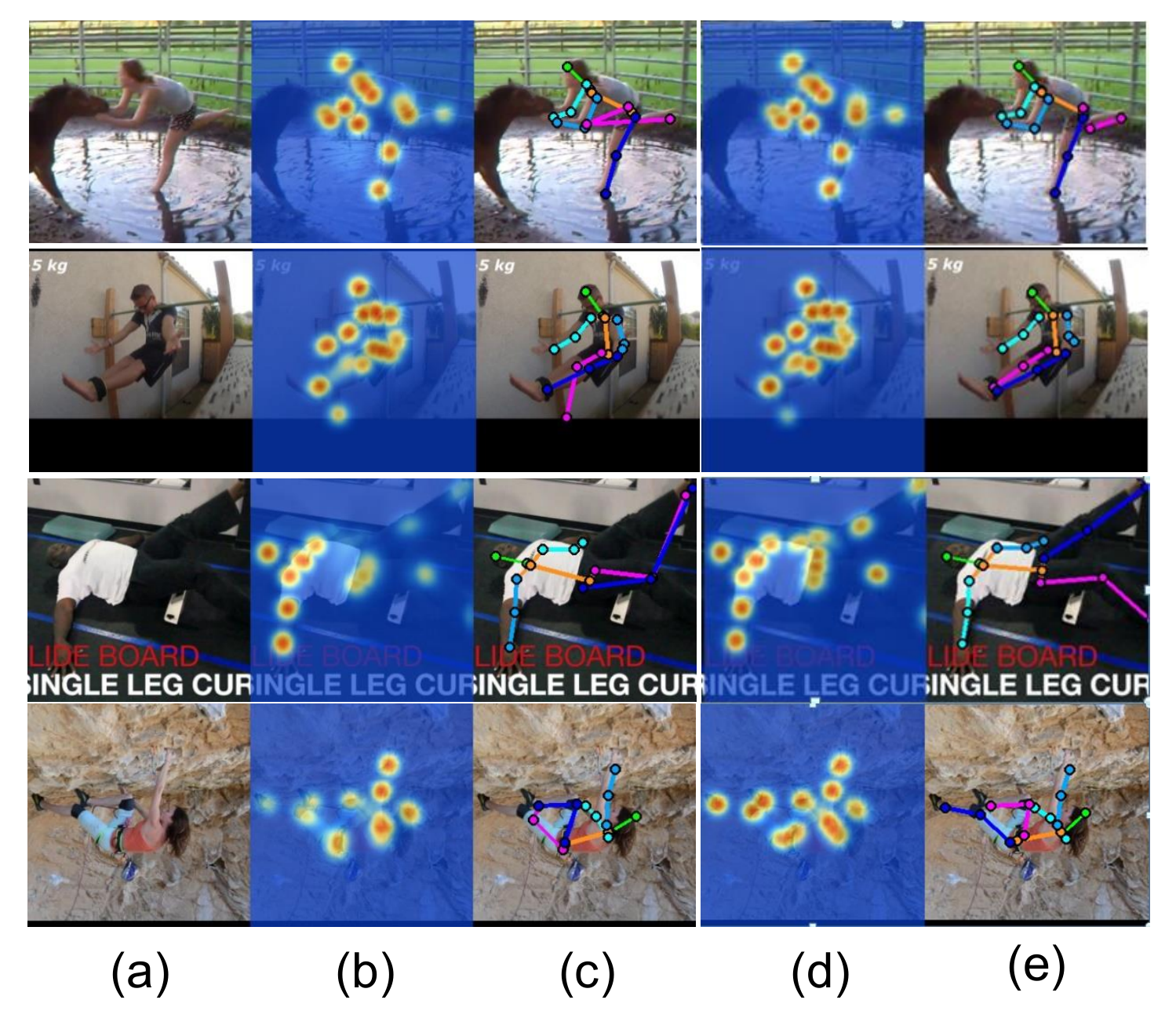}
\end{center}
\caption{ Qualitative evaluation on MPII Human pose test dataset. (a) The input images. (b) Heatmaps predicted by the original hourglass network. (c)Skeletons predicted by the original hourglass network. (d) Further refined heatmaps predicted by our proposed network. (e) Further refined skeletons produced by our proposed network.}
\label{fig:image5}
\end{figure}

\begin{figure}
\begin{center}
\includegraphics[height=0.4\columnwidth,width=1.0\columnwidth]{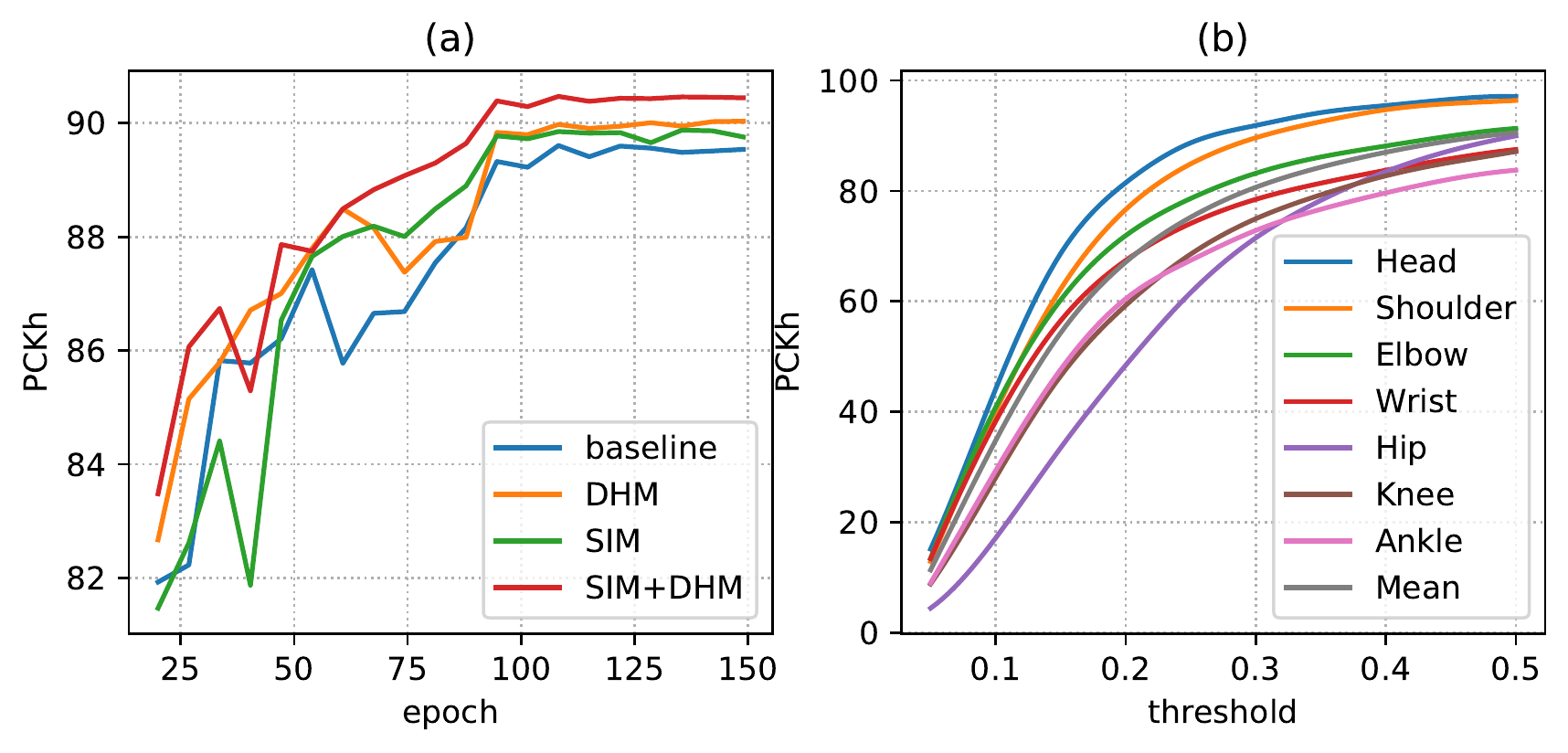}
\end{center}
\caption{(a) PCKh curves of the training process across different networks on MPII validation set. (b) PCKh curves of different keypoints predicted by SPCNet at threshold of 0.05 to 0.5 on MPII validation set.}
\label{fig:image7}
\end{figure}

\noindent{\bf Comprehensive Analysis.} In this experiment, we explore the contribution of each module to the whole network. Besides separately adding each proposed module to the baseline, we further employ the two proposed modules to the baseline simultaneously. The results are reported in Table\ref{tab:each}. Compared with the 88.9\% PCKh score of the baseline hourglass network, we achieve 0.8\% improvement with only the \textit{Dilated Hourglass Module} used and 0.7\% improvement with only the \textit{Selective Information Module} used. Finally our method achieves 90.0\% PCKh score with the two proposed modules applied simultaneously, which is 1.1\% improvement compared to the baseline hourglass. Validation PCKh curves across different architectures above and the validation PCKh curves of different keypoints at different threshold for SPCNet are plotted in Figure \ref{fig:image7}(a) and (b) respectively.

\begin{figure*}
\begin{center}
\includegraphics[height=5cm, width=16cm]{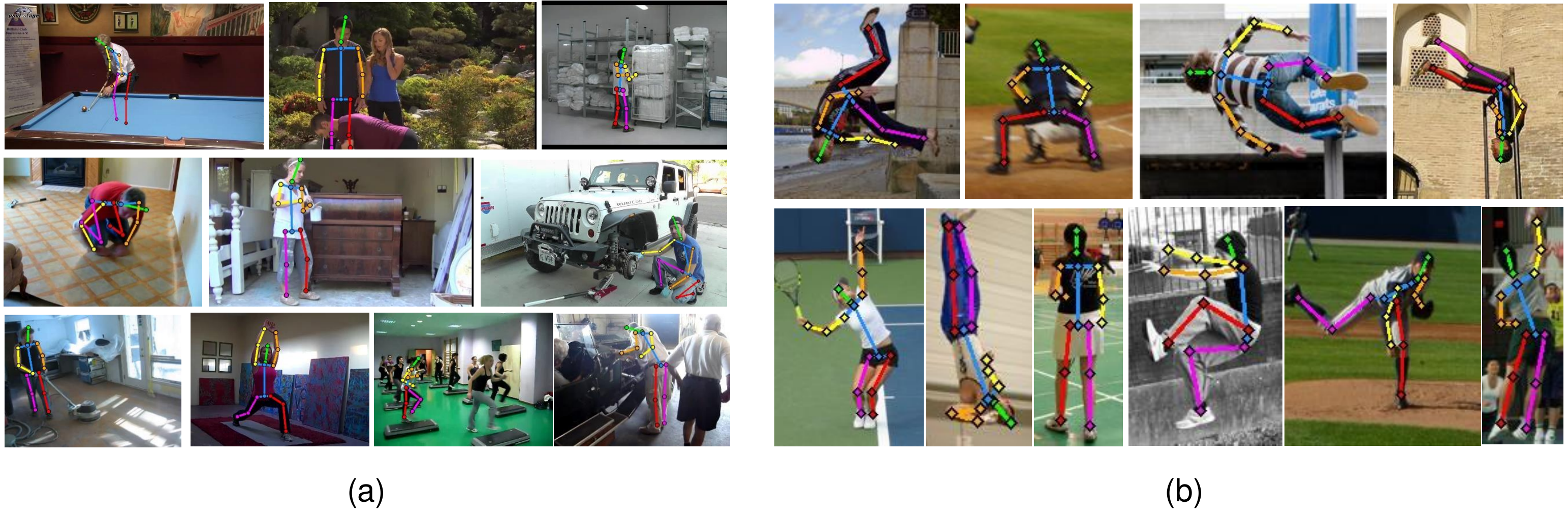}
\end{center}
\caption{(a) Examples of estimated poses on the MPII Human Pose test dataset. (b) Examples of estimated poses on the LSP test dataset.  Our model deals well with occlusions, blurred and twisted limbs, abnormal poses, and changes of view position.}
\label{fig:image6}
\end{figure*}

\begin{table}
\begin{center}
\caption{Comparisons of score on the MPII Human Pose test dataset (PCKh@0.5).}
\label{tab:mpii}
\resizebox{1.0\columnwidth}{!}{
\begin{tabular}{lllllllll}
\hline
  Methods       &  Head & Sho. & Elb. & Wri. & Hip & Knee& Ank.& Total  \\
\hline
Lifshitz et al.\cite{lifshitz2016human}	&97.8&93.3&85.7&80.4&85.3&76.6&70.2&85.0\\
Rafi et al. \cite{rafi2016efficient} &97.2&93.9&86.4&81.3&86.8&80.6&73.4&86.3\\
Insafutdinov et. al.\cite{insafutdinov2016deepercut}	&96.8&95.2&89.3&84.4&88.4&83.4&78.0&88.5  \\
Wei et al.\cite{wei2016convolutional} &97.8&95.0&88.7&84.0&88.4&82.8&79.4&88.5 \\
Bulat$\&$Tzimiropoulos \cite{bulat2016human}	&97.9&95.1&89.9&85.3&89.4&85.7&81.7&89.7 \\
Newell et al.\cite{newell2016stacked}  &98.2 &96.3& 91.2& 87.1& 90.1& 87.4& 83.6& 90.9    \\
Sun et al.\cite{sun2017human} &98.1 &96.2& 91.2 &87.2& 89.8& 87.4& 84.1& 91.0  \\
Chu et al.\cite{chu2017multi} &98.5 &96.3 &91.9 &88.1 &90.6 &88.0 &85.0 &91.5  \\
Chen et al.\cite{chen2017rethinking}	&98.1&96.5&	92.5&88.5&90.2&89.6&86.0&91.9\\
Yang et al.\cite{yang2017learning}	&98.5&96.7&92.5	&88.7&91.1&88.6&86.0&92.0 \\
Ke et al.\cite{ke2018multi} &98.5 &96.8 &92.7& 88.4& 90.6& 89.3& 86.3& 92.1  \\
Tang et al.\cite{tang2018deeply}	&98.4&96.9&92.6&88.7&91.8&89.4&86.2&92.3\\
HRnet \cite{Sun2019Deep} &98.6& 96.9& 92.8& {\bf89.0} &91.5 &89.0& 85.7& 92.3\\
Zhang et al.\cite{zhang2019human} &98.6& 97.0& 92.8& 88.8 & 91.7& {\bf89.8}&{\bf86.6} & 92.5\\
\hline
\textbf{Ours(SPCNet)} & {\bf 98.8} & {\bf97.1}& {\bf 93.2}& 88.9& {\bf92.0}& 89.6& 86.3 & {\bf92.6}   \\
\hline
\end{tabular}
}
\end{center}
\end{table}

Following prior methods\cite{chu2017multi,yang2017learning}, we further conduct the horizon flip and six-scale image pyramids on the validation set of the MPII Human Pose. The results are reported in Table \ref{tab:six}. With the multi-scale image pyramids with horizon flipping applied to the prediction, we achieve 91.1\% PCkh score, which is 0.3\% improvement compared to the HRNet~\cite{Sun2019Deep}.

\subsection{Comparison with the state-of-the-art Methods}\label{subsection3}

To evaluate the performance of our method, we compare our network with the prior state-of-the-art methods on three datasets: MPII Human Pose test dataset, LSP test dataset and FLIC test dataset. Moreover, we give some qualitative results generated by baseline network and our proposed network. 

\noindent{\bf MPII Human Pose dataset.} We report the PCKh scores of our approach and the previous state-of-the-art methods at the threshold of 0.5 in Table \ref{tab:mpii}. Compared with the hourglass network, our approach improves the performance of total PCKh score from 90.9\% to 92.6\%.  Specifically, as shown in Table \ref{tab:mpii}, our method surpasses the HRNet\cite{Sun2019Deep} across all keypoints except for the wrist. The final results demonstrate the superior performance of our proposed model over the prior state-of-the-art methods in terms of PCKh score.

\noindent{\bf LSP dataset.} Table \ref{tab:lsp} summarizes the PCK scores at the threshold of 0.2 on LSP dataset. We follow the previous methods\cite{yang2017learning,chu2017multi} to train our network by adding training set of MPII Human Pose to the LSP and its extend training set. Our method achieves the highest total score 96.4\% and exceeds the previous state-of-the-art results across all keypoints on the LSP test set. We observe that the proposed network improves the PCK scores with a large margin by 4.4\% and 2.7\% on the wrist and elbow compared with the closest competitor, and obtains 1.3\% improvement in average.

\begin{table}
\begin{center}
\caption{Comparisons of score on the LSP test dataset (PCK@0.2).}
\label{tab:lsp}
\resizebox{1.0\columnwidth}{!}{
\begin{tabular}{lllllllll}  
\hline
  Methods      & Head   & Sho. & Elb.& Wri.& Hip& Knee& Ank.& Total\\
\hline
Belagiannis$\&$Zisserman \cite{belagiannis2017recurrent} &95.2 &89.0 &81.5 &77.0& 83.7 &87.0 &82.8& 85.2\\
Lifshitz et al.\cite{lifshitz2016human}  &96.8 &89.0 &82.7 &79.1& 90.9& 86.0& 82.5& 86.7\\
Pishchulin et al.\cite{pishchulin2016deepcut}  &97.0&91.0&83.8&78.1&91.0&86.7&82.0&87.1\\

Insafutdinov et al.\cite{insafutdinov2016deepercut}& 96.8  & 95.2 & 89.3 & 84.4 &88.4 & 83.4& 78.0 & 88.5 \\
Wei et al.\cite{wei2016convolutional} &97.8 &92.5& 87.0 &83.9 &91.5 &90.8 &89.9 &90.5 \\
Bulat $\&$Tzimiropoulos\cite{bulat2016human}& 97.2  & 92.1 & 88.1 & 85.2 &92.2 & 91.4& 88.7 & 90.7 \\
Sun et al.\cite{sun2017human} & 97.9 & 93.6 & 89.0 & 85.8 & 92.9 & 91.2 & 90.5 & 91.6     \\
Chu et al.\cite{chu2017multi}& 98.1  & 93.7 & 89.3 & 86.9 &93.4 & 94.0& 92.5 & 92.6    \\
Yang et al.\cite{yang2017learning}  &98.3 & 94.5 & 92.2 & 88.9 & 94.4 & 95.0 & 93.7 & 93.9   \\
Zhang et al.\cite{zhang2019human} &{\bf98.4} &94.8 &92.0 &89.4 &94.4 &94.8 &93.8 &94.0 \\
Tang et al.\cite{tang2018deeply} & 98.3 & 95.9 & 93.5 & 90.7 & 95.0 & 96.6 & 95.7 & 95.1   \\
\hline
\textbf{Ours(SPCNet)}  &98.3 & {\bf96.3} &{\bf96.2} & {\bf95.1}  & {\bf96.0} &  {\bf96.7} &  {\bf95.9} & {\bf96.4}\\
\hline
\end{tabular}
}
\end{center}
\end{table}

\noindent{\bf FLIC dataset.} Table \ref{tab:FLIC} shows the PCK@0.2 scores on FLIC dataset. Our proposed method achieves the 99.3\% and 98.2\% PCK@0.2 scores for the elbow and wrist, which are 0.3\% and 1.2\% improvement compared with the hourglass network respectively.

\begin{table}
\begin{center}
\caption{Comparisons of score on the FLIC test dataset (PCK@0.2).}
\label{tab:FLIC}
\resizebox{0.6\columnwidth}{!}{
\begin{tabular}{lccccccccc}
\hline
  Methods     & Elbow & Wrist & Total\\
\hline
Wei et al.\cite{wei2016convolutional} &97.8& 95.0& 96.4& \\
Newell et al.\cite{newell2016stacked}  &99.0 &97.0&  98.0&   \\
Ke et al.\cite{ke2018multi} &99.2 & 97.3&98.3&\\
\hline
\textbf{Ours(SPCNet)}  &{\bf99.3} & {\bf98.3} &{\bf98.8} &\\
\hline
\end{tabular}
}
\end{center}
\end{table}

\noindent{\bf Qualitative results.} We compare the baseline model (8-stack hourglass network) with our proposed model by visualizing the estimated heatmaps and skeletons on the test set of MPII Human Pose, as demonstrated in Figure \ref{fig:image5}. We observe that our method outperforms the baseline model in the challenging cases, such as occluded keypoints, invisible keypoints, crowded background and abnormal body posture. The 1st row shows a women with a twisted pose. The 2nd row presents person with overlapped limbs. Then the 3rd row exhibits a person whose limbs are indistinct with surroundings, and the 4th row shows a person with abnormal pose. The baseline model produces the failure predictions as shown in Figure \ref{fig:image5}(c) while our proposed model makes the refined predictions as shown in Figure \ref{fig:image5}(e) when facing these kinds of complex scenarios. It is noteworthy that our proposed model generated higher response in the prediction heatmaps with the blurred regions, as illustrated in Figure \ref{fig:image5}(d) compared to the Figure \ref{fig:image5}(b). In addition, examples of estimated pose on MPII test dataset and LSP test dataset are illustrated in Figure \ref{fig:image6}.

Compared with the baseline hourglass, our method can effectively improve the pose estimation performance of difficult scenarios (e.g., occlusion, twisted and overlapped human body, abnormal pose, and so on). By using the ~\textit{Dilated Hourglass Module}, we can preserve the spatial resolution information along with large receptive field. With the ~\textit{Selection Information Module}, the multi-stage and multi-scale information can be adaptively selected and enhanced for the final human keypoints localization. Leveraging the two proposed modules, we can get high spatial resolution and different receptive field information, which are critical to the human pose estimation task.

\section{Conclusion}
\label{sec5}
In this paper, we propose to incorporate a \textit{Dilated Hourglass Module} and a \textit{Selective Information Module} into an end-to-end architecture for human pose estimation. By stacking the \textit{Dilated Hourglass Module}, we can preserve spatial resolution information along with large receptive field. Meanwhile, a \textit{Selective Information Module} is designed to select relatively important features from different levels under a sufficient consideration of spatial content-aware mechanism. The effectiveness of the \textit{Dilated Hourglass Module} and the \textit{Selective Information Module} are evaluated on validation set of MPII Human Pose. We experimentally observe that the proposed network can alleviate the difficulties brought by occlusions, overlapped bodies and abnormal poses. Overall, our approach  achieves state-of-the-art results on MPII Human Pose dataset, LSP dataset and FLIC dataset.

\bibliography{ecai}
\end{document}